\documentclass[letterpaper]{article}
\usepackage{aaai2026}  % DO NOT CHANGE THIS
\usepackage{times}  % DO NOT CHANGE THIS
\usepackage{helvet}  % DO NOT CHANGE THIS
\usepackage{courier}  % DO NOT CHANGE THIS
\usepackage[hyphens]{url}  % DO NOT CHANGE THIS
\usepackage{graphicx} % DO NOT CHANGE THIS
\usepackage{natbib}  % DO NOT CHANGE THIS AND DO NOT ADD ANY OPTIONS TO IT
\usepackage{caption} % DO NOT CHANGE THIS AND DO NOT ADD ANY OPTIONS TO IT
\usepackage{algorithm}
\usepackage{algorithmic}
\usepackage{amsmath}
\usepackage{amssymb}
\usepackage{multirow}
\usepackage{times}
\usepackage{helvet}
\usepackage{courier}
\usepackage{xcolor}
\usepackage{dsfont}

\usepackage{newfloat}
\usepackage{listings}
\DeclareCaptionStyle{ruled}{labelfont=normalfont,labelsep=colon,strut=off} % DO NOT CHANGE THIS
\floatstyle{ruled}
\newfloat{listing}{tb}{lst}{}
\floatname{listing}{Listing}
\title{SAMIR, an efficient registration framework via robust feature learning from SAM}
\author{
    Yue He\textsuperscript{\rm 1}\equalcontrib, Min Liu\textsuperscript{\rm 1}\equalcontrib, Qinghao Liu\textsuperscript{\rm 1}, Jiazheng Wang\textsuperscript{\rm 1}, Yaonan Wang\textsuperscript{\rm 1}, Hang Zhang\textsuperscript{\rm 1}, Xiang Chen\textsuperscript{\rm 1} \thanks{Corresponding authors: xiangc@hnu.edu.cn}
}
\affiliations{
    \textsuperscript{\rm 1}College of Electrical and Information Engineering, Hunan University, Changsha, Hunan, China
}

\usepackage{bibentry}
\makeatletter
\begin{document}

\maketitle

\begin{abstract}
Image registration is a fundamental task in medical image analysis. Deformations are often closely related to the morphological characteristics of tissues, making accurate feature extraction crucial. Recent weakly supervised methods improve registration by incorporating anatomical priors such as segmentation masks or landmarks, either as inputs or in the loss function. However, such weak labels are often not readily available, limiting their practical use. Motivated by the strong representation learning ability of visual foundation models, this paper introduces SAMIR, an efficient medical image registration framework that utilizes the Segment Anything Model (SAM) to enhance feature extraction. SAM is pretrained on large-scale natural image datasets and can learn robust, general-purpose visual representations. Rather than using raw input images, we design a task-specific adaptation pipeline using SAM's image encoder to extract structure-aware feature embeddings, enabling more accurate modeling of anatomical consistency and deformation patterns. We further design a lightweight 3D head to refine features within the embedding space, adapting to local deformations in medical images. Additionally, we introduce a Hierarchical Feature Consistency Loss to guide coarse-to-fine feature matching and improve anatomical alignment. Extensive experiments demonstrate that SAMIR significantly outperforms state-of-the-art methods on benchmark datasets for both intra-subject cardiac image registration and inter-subject abdomen CT image registration, achieving performance improvements of 2.68\% on ACDC and 6.44\% on the abdomen dataset. The source code will be publicly available on GitHub following the acceptance of this paper.
\end{abstract}

\section{Introduction}

Pair-wise image registration is essential in medical image processing, aligning moving and fixed images from different subjects, modalities, time points, or perspectives to enhance diagnosis, surgical planning, and motion analysis \cite{chen2021deepSurvey}.

Traditional registration methods rely on iterative optimization to minimize the distance like normalized cross-correlation (NCC) or mean square error (MSE), between warped moving and fixed images, which is time-consuming. Deep learning-based registration has significantly accelerated this process, enabling registration through a single forward inference after training. However, these methods typically employ conventional loss functions designed to minimize raw image contrast discrepancies. Crucially, they fail to take into account inherent variations in image quality (e.g., contrast heterogeneity and noise patterns) that arise from differences in imaging equipment, acquisition protocols, and operator techniques, even for repeated scans of the same subject.
%However, these methods typically adopt loss functions similar to those used in traditional approaches, focusing on minimizing the dissimilarity between raw image contrasts. They do not account for the influence of image quality, which may vary across different imaging equipment, different time and different operators, even for the same subject, in terms of different contrast and potential noises.
% However, these methods often use similar loss functions as traditional approaches, minimizing dissimilarity between raw image contrasts.  This approach assumes consistent lighting and noise-free conditions, leading to suboptimal performance in real-world scenarios.

\begin{figure}[t]
\centering
\includegraphics[width=1\columnwidth]{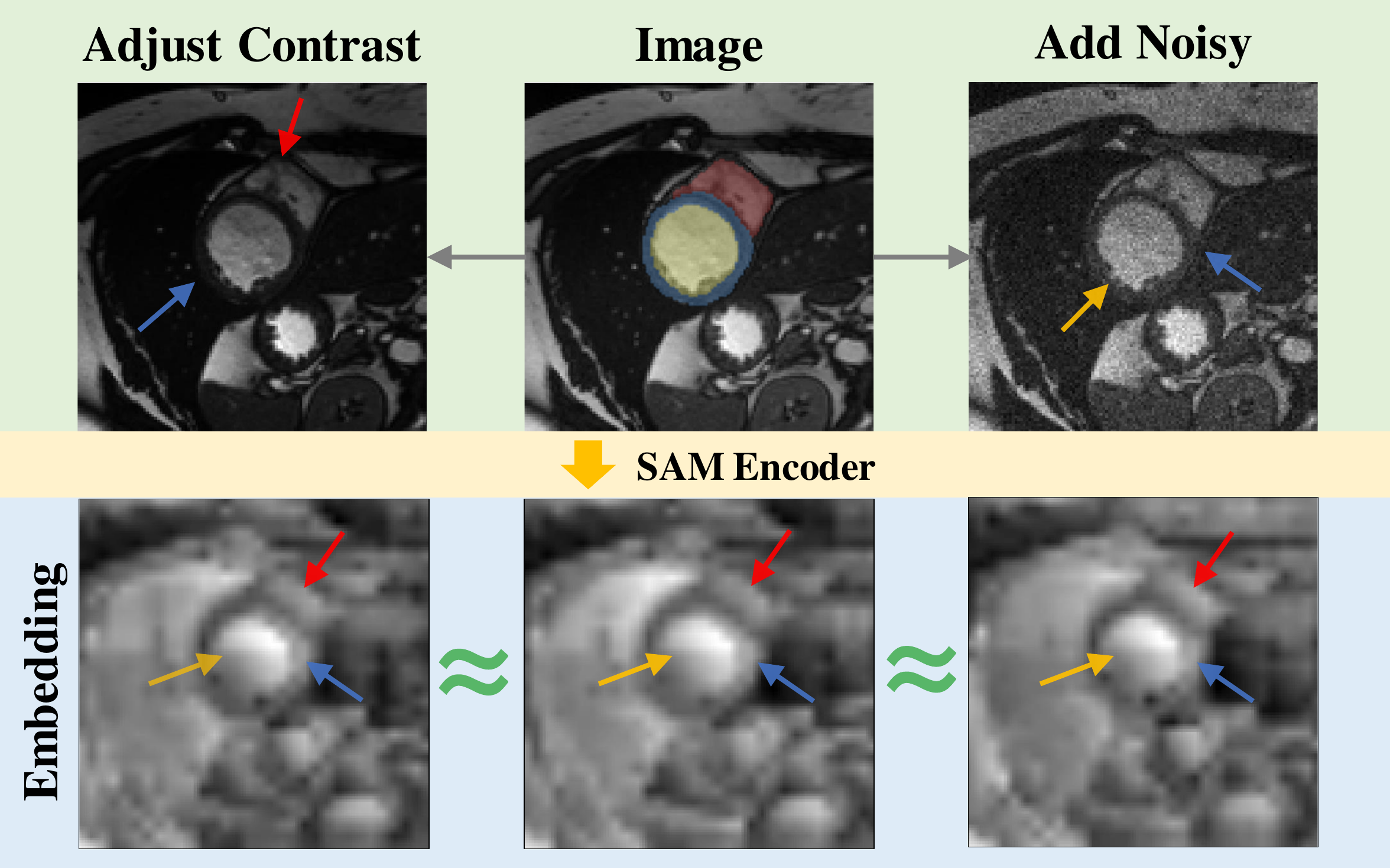} 
\caption{Feature extraction from SAM encoder for different scenarios of images. It can be observed that the embeddings of the SAM image encoder show strong robustness, accurately capturing object structures even with contrast changes and noise interference.}
\label{fig:intro}
\end{figure}

Current deep learning methods enhance feature learning in registration networks through various architectures, either explicitly or implicitly. Early studies used convolutional layers for feature extraction and integration \cite{balakrishnan2019voxelmorph,dalca2019varreg},  and later incorporated transformers to capture long-range correlations \cite{chen2022transmorph}. Following research have also explored the large kernel convolution \cite{jia2022u} and multi-layer perceptron \cite{meng2024correlation} to further enhance the feature learning in the registration network. Some approaches also proposed to learn explicit representations from the input images and then fed them into the registration block \cite{lee2019image}. For example, Lee et al. \cite{lee2019image} proposed an image-and-spatial transformer network to leverage the structure of information to learn new image representations that are optimized for the downstream registration task. Previous research has demonstrated that incorporating anatomical structure priors like segmentation masks or landmarks in network training can lead to significantly better registration performance \cite{balakrishnan2019voxelmorph,chen2021deepDiscontinuity,superwarp}. These approaches, also known as the weakly-supervised registration methods, either utilize segmentation/landmarks as a loss function in the network training \cite{balakrishnan2019voxelmorph}, or directly incorporate them as the input of the network \cite{chen2021deepDiscontinuity}. However, weak labels such as segmentation masks are not always available, limiting their practical applications.

Recently, the Segment Anything Model (SAM)\cite{sam} has significantly revolutionized the field of image segmentation. One of its key strength lies in generalization capabilities, enabled by training on a diverse dataset with extreme lighting, complex occlusion, and boundary uncertainties. This allows SAM to achieve impressive zero-shot performance across tasks, often matching or surpassing fully supervised results. In addition, the feature embeddings learning from the SAM encoder exhibit strong robustness to either contrast variations or noise, as shown in Figure \ref{fig:intro}. % In addition, this grants SAM strong robustness as shown in Figure \ref{fig:intro}.
% , for cardiac MRI slices, the structural features of key tissues are well preserved, whether the image contrast is altered or noise is added. 
Fine-tuning SAM on medical datasets has further enhanced its segmentation capabilities\cite{sammed2d,MedSAM,yue_surgicalsam,Lei2025medlam}. While SAM has achieved success in numerous downstream tasks, its application in the field of image registration remains limited. 
% To the best of our knowledge, only Huang et al\cite{samreg} has proposed SAMReg, which formulates image registration as the task of identifying corresponding regions of interest between moving and fixed images, leveraging SAM for multi-class segmentation on image pairs. This underscores the potential for further exploration in this area.

To the best of our knowledge, there are currently only two works that apply SAM to medical image registration. 
One approach is SAMReg \cite{samreg}, which leveraged SAM for multi-class segmentation of image pairs instead of directly predicting a global deformation field, aiming to achieve accurate local registration.
Another is SAM-Assisted Registration \cite{Xu_2025_IPMI}, which utilized text prompts to guide SAM in generating segmentation masks and then employed them to assist both training and inference, ensuring anatomical consistency. 
While they do not directly utilize SAM in the deformation fields prediction, their efforts demonstrate the promising potential for further exploration in this field.

% To the best of our knowledge, there are currently only two works that apply SAM to medical image registration. SAMReg \cite{samreg} formulates image registration as the task of identifying corresponding regions of interest between the moving and fixed images. This method leverages SAM for multi-class segmentation of image pairs, rather than directly predicting a global deformation field, with the aim of achieving accurate local registration.
% The other is a novel SAM-assisted registration framework named SAM-Assisted Registration \cite{Xu_2025_IPMI}, which combines prototype learning with contour-aware mechanisms. This approach used text prompts to guide SAM in generating segmentation masks, which are then used to assist both training and inference, ensuring anatomical consistency. However, the model architecture is relatively complex.
% These early efforts highlight the promising potential for further exploration in this field.

% In this paper, we propose a novel registration framework \textbf{SAMIR} (\textbf{S}egment \textbf{A}nything \textbf{M}odel for \textbf{I}mage \textbf{R}egistration) based on feature learning from SAM to approach robust and more accurate image registration. SAMIR is designed to tackle both robust feature learning and large deformation in the network, incorporated with a hybrid loss to aliogh the moving and fixed images in multi-space. 
% To the best of our knowledge, our work represents the first attempt to apply SAM to traditional registration subtasks, opening new avenues for research and development.

In this paper, we propose \textbf{SAMIR} (\textbf{S}egment \textbf{A}nything \textbf{M}odel for \textbf{I}mage \textbf{R}egistration), a novel medical image registration framework that leverages SAM to enhance feature learning, aiming to achieve more robust and accurate image alignment. SAMIR introduces a registration-specific adaptation pipeline for SAM, utilizing its powerful pretrained image encoder and complementing it with a lightweight 3D head to extract accurate, fine-grained, and robust image features. Furthermore, a hybrid loss function is introduced to align the moving and fixed images across multi-scale spaces, addressing the challenges of robust feature learning and large deformation modeling.

In summary, the contributions of this paper can be summarized as follows: 
\begin{itemize}
\item We proposed a novel feature-driven registration framework, SAMIR, leveraging the structure-aware properties of foundation models to achieve robust and accurate medical image registration.
\item A novel feature-level loss is designed to further enhance the structure alignment and robustness on registration.
\item Our SAMIR achieves state-of-the-art (SOTA) performance on multiple registration tasks, including the ACDC dataset and abdomen dataset, with Dice scores improved by 2.68\% and 6.44\%, respectively.
\end{itemize}

\section{Related Works}

\subsection{Deformable Medical Image Registration}
Early image registration networks typically employ a single-step approach for deformation field estimation. For example, VoxelMorph\cite{balakrishnan2019voxelmorph} adopts an end-to-end U-Net architecture to predict the deformation fields. Such types of methods can handle small local deformation well, while they often struggle with large deformations. Following research like TransMorph\cite{chen2022transmorph} and LKU-Net\cite{jia2022u} proposed to replace convolutional modules with Transformer components and large kernel convolution, to enlarge the perceptual fields.
% While obtaining better registration accuracy, the registration for large deformation remains unsolved.
Despite improved accuracy, large deformation registration remains challenging.
Recent work has shifted toward more sophisticated architectures to improve accuracy and robustness. The main approaches fall into two categories. The first category is adapting cascaded structures. For instance, VTN \cite{zhao2019recursive}, which utilized recursive cascaded networks, splitting the registration task into a multi-step cascade small deformation. VR-Net \cite{vrnet} splits image registration into closed-form and denoising subproblems, models them with specialized layers, and cascades them for fast, accurate, data-efficient registration. The second category is based on coarse-to-fine pyramid strategies. For instance, LapIRN\cite{mok2020large} introduced a deep Laplacian pyramid network to progressively refine deformations from low to high resolution. RPD\cite{wang_RDP} integrated recursion into the pyramid framework to better handle large displacements, while CorrMLP\cite{meng2024correlation} presented the first MLP-based coarse-to-fine registration model. These strategies enhance multi-scale modeling and substantially enhance complex registration performance.
% significantly improve performance in complex registration tasks.

\subsection{Visual Foundation Models}
Foundation models have recently seen rapid development in computer vision, achieving strong generalization capabilities and good adaptability to a wide range of downstream tasks through pre-training on large-scale and diverse datasets. Segment Anything (SAM)\cite{sam} demonstrated impressive zero-shot performance across various image segmentation tasks via pre-training on over one billion masks. SegGPT\cite{seggpt} unified image segmentation into a general visual perception task by transforming diverse segmentation tasks into identically formatted images and incorporating a random color mapping mechanism for contextual coloring, thereby establishing a context-based learning framework for universal segmentation. DINO\cite{dino} addressed unsupervised visual representation learning via self-distillation and contrastive learning with a student-teacher architecture and pseudo-labeling, enabling transferable feature learning from large-scale unlabeled data and achieving strong performance. DINOv2\cite{dinov2} combined DINO and iBOT\cite{ibot} with KoLeo regularization and Sinkhorn-Knopp centering to learn robust visual features without supervision, achieving strong performance on classification, segmentation, and retrieval tasks. 

% Despite the remarkable success of foundation models across various domains, their application to medical image registration remains largely unexplored. 
% On one hand, image registration is intrinsically challenging, involving multi-modality and large deformations. 
While foundation models have shown considerable success in computer vision, their application to medical image registration remains challenging due to: (1)the lack of reliable ground-truth deformation fields for training and validation, and (2) the scarcity of publicly available, well-paired medical image datasets, further complicated by their large volumetric dimensions.
% On the one hand, image registration lacks reliable ground-truth deformation fields. 
% On the other hand, the large volume and high resolution of 3D images impose higher demands on computational resources.
Notably, existing research \cite{DONG201754,jointsegmentationdiscontinuitypreservingdeformable} has demonstrated that registration and segmentation share core feature-matching principles, enabling mutual reinforcement.
Capitalizing on this theoretical connection, we propose SAMIR, a novel framework that leverages SAM's structure-aware properties to guide deformation field estimation in registration tasks.
%Therefore, although general visual foundation models have made significant progress, the construction of large models tailored for accurate and robust medical image registration remains in its early exploration stage.

\section{Method}
% Pair-wise image registration establishes spatial correspondence between the moving image $\textbf{I}_M$ and fixed image $\textbf{I}_F$, and the deformation function $\phi(\circ)$ from $\textbf{I}_M$ to $\textbf{I}_F$ is formulated as, 
Given paired moving image $I_m$ and fixed image $I_f$ defined over the spatial domain $\Omega \subseteq {R}^3$, image registration aims to estimate a deformation field $\phi(\circ): {R}^3 \to {R}^3$ that optimally aligns $I_m$ to $I_f$, such that $I_m \circ \varphi \approx I_f$. The deformation field is typically formulated as,
\begin{equation}
\label{eqn:formula}
\phi(\textbf{x}) = \textbf{x} + u(\textbf{x}),
\end{equation}
where, $\textbf{x}$ represents voxel coordinates, and $u(\textbf{x})$ denotes the displacement field.

\begin{figure*}[htb]
\begin{center}
\includegraphics[width=1\textwidth]{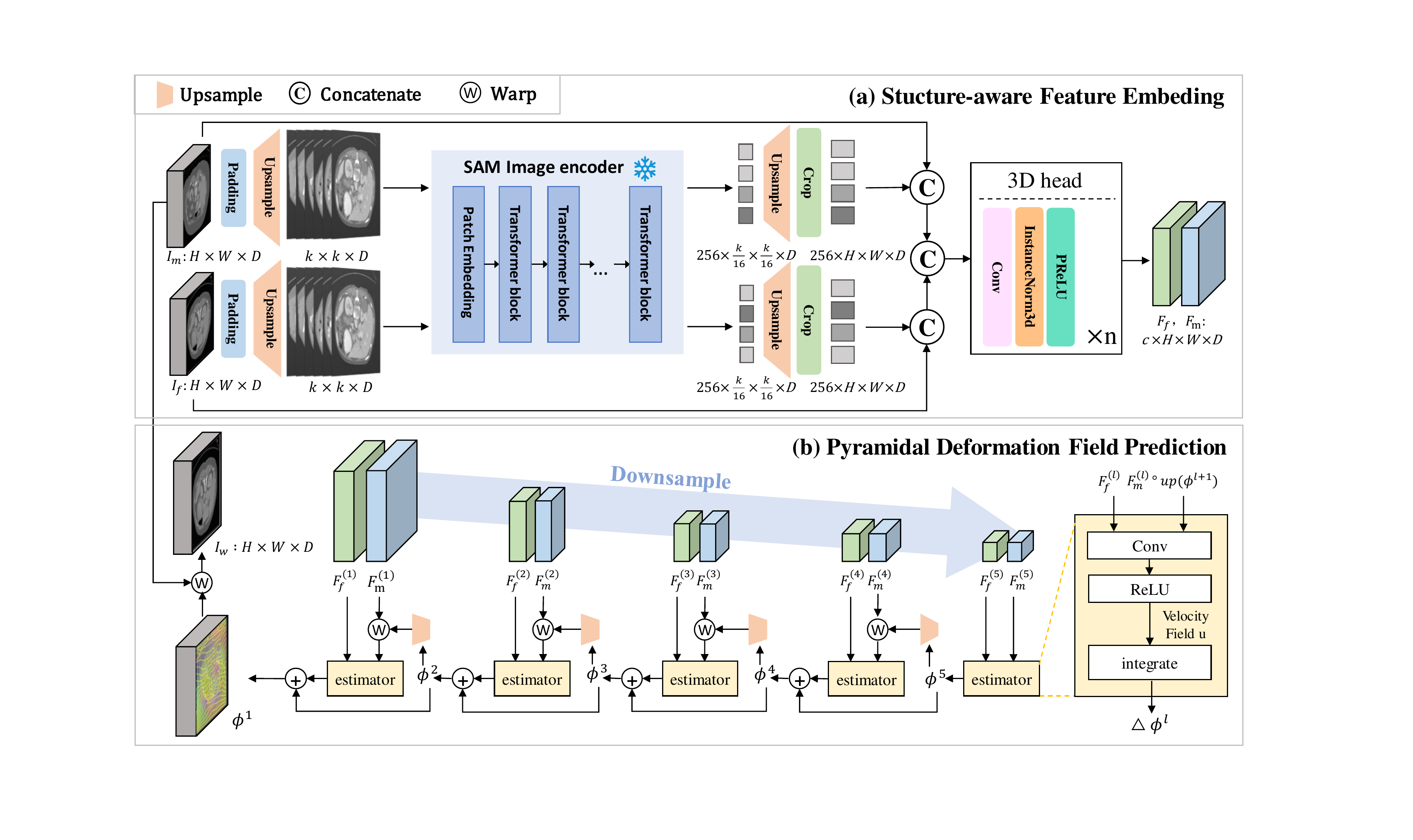}
    \caption{Overview of SAMIR Network Architecture. The SAMIR network leverages the SAM pre-trained image encoder to extract stucture-aware feature embeddings and employs a pyramidal deformation field prediction block to achieve coarse-to-fine progressive displacement field optimization, ultimately accomplishing accurate cross-modality medical image registration.}
    \label{fig:samir}
\end{center}
\end{figure*}

% As shown in Fig. \ref{fig:samir}, the network architecture of the SAMIR can be primarily divided into two parts: 1) A Structure-Aware Feature Embedding (SAFE) block to learn robust structure representation from $\textbf{I}_M$ and $\textbf{I}_F$ via the foundation model SAM pre-trained on large scale images.  2) A Pyramidal Deformation Field Prediction (PDFP) block that adopts a coarse-to-fine progressive optimization strategy to generate high-precision displacement fields, covering both small and large deformation tasks. Note that, different from SAMReg \cite{samreg} utilizing SAM for segmenting structure of interest, we only utilize the latent feature embedding of SAM for downstream tasks.

As illustrated in Figure \ref{fig:samir}, the SAMIR framework comprises two principal components: (1) A Structure-Aware Feature Embedding (SAFE) module that extracts robust structural intermediate representations from both $I_m$ and $I_f$ using the encoder from a visual foundation model, subsequently performing feature integration and medical domain adaptation via a 3D convolutional head; (2) A Pyramid Deformation Field Prediction (PDFP) module employing a coarse-to-fine progressive optimization strategy to generate high-precision displacement fields adaptable to various deformation magnitudes. Notably, the foundation model encoder remains frozen throughout training, substantially reducing trainable parameters while conserving computational resources and training time.

\subsection{Stucture-Aware Feature Embeding} 
In this paper, we utilize SAM as the foundation model of our SAMIR for feature encoding. The SAM image encoder is a Vision Transformer (ViT) pretrained via Masked autoencoders \cite{MaskedAutoencoders2021} on a large and diverse dataset, endowing it with robust structure-aware properties and powerful zero-shot transfer performance.
% strong generalization capabilities

\textbf{Dimensional Adaptation of Foundation Models.}
To adapt the fundamental vision model for medical image registration, we address several critical discrepancies between standard natural images and medical imaging data. (1)Current foundation models such as SAM are predominantly designed for 2D natural images, whereas medical imaging largely operates in 3D space. (2)While SAM requires square input dimensions $(H = W)$, medical images frequently exhibit arbitrary aspect ratios. (3) A significant resolution gap exists between SAM's training data (shorter side resized to 1500 pixels) and medical images (generally lower resolutions such as $128 \times 128$).
% typically 3300 $\times$ 4950 pixels
To bridge these gaps, we implement a multi-stage adaptation strategy. For 3D compatibility, we follow \cite{masam}, processing volumetric medical data $[B, H, W, D]$ as sequential 2D slices $[BD, H, W]$, extracting features slice-wise $F_{slice}= \left \{ f^{1},f^{2},\dots ,f^{D} \right \} ,f^{i} \in R^{C\times H \times W} $, and then reorganizing them into coherent 3D embeddings $F_{ori}$. This approach effectively leverages the 2D backbone while preserving volumetric information. 
For handling varying aspect ratios, we employ strategic padding from $[H,W,D]$ to $[H',W',D]$ to conform to SAM's square input requirements, recording padding parameters to enable precise restoration to original dimensions during post-processing.
Resolution mismatches are mitigated through intelligent upsampling of medical images from $[H',W',D]$ to $[k,k,D]$ to match the encoder's expected input dimensions prior to feature extraction.

In addition, the feature embeddings of the SAM encoder are at 1/16th the spatial resolution of the original input, which leads to loss of spatial information and inaccurate prediction of deformation fields. To tackle this issue, we apply a two-stage upsample operation (the inputs are upsampled to $(k/H') \times$ prior to the encoder, and the features are upsampled $(16H/k) \times$ ) to upsample the obtained feature embeddings to match the input size, resulting in a dimension of $[256, H, W, D]$.
% \begin{align}
%     & I_{input} = up_2\left ( Pad\left ( I^{i},\left ( H',W' \right )  \right ),(k,k)\right ) &\\
%     & F_{ori} = Stack\left ( SAM\_encoder\left ( I_{input}^{i}  \right )  \right ) &\\
%     & F_{gen} = Crop(up_2\left ( F_{ori},\left ( C,k,k,D \right )  \right ) ,\left ( C,H,W,D \right )) &
% \end{align}

\begin{algorithm}[tb]
\caption{registration-specific Adaptation Pipeline}
\label{alg:adaptation_pipeline}
\textbf{Input}: Image ${I} \in {R}^{H \times W \times D}$ \\
\textbf{Parameter}: Padding size $(H', W')$, Target size $k$ \\
\textbf{Output}: Feature embeddings $F_{\text{final}} $
% \in {R}^{B \times C \times H \times W \times D}
\begin{algorithmic}[1]
\STATE \textbf{Preprocess Input Slices}
\FOR{$i = 1$ to $D$}
    \STATE $I^i \gets \text{Slice}({I}, i)$ \COMMENT{Extract $i$-th slice}
    \STATE $\hat{I}^i \gets \text{Pad}(I^i, (H', W'))$
    \STATE $I_{\text{input}}^i \gets \text{Up}(\hat{I}^i, (k, k))$
\ENDFOR

\STATE \textbf{Extract and Stack Features}
\FOR{$i = 1$ to $D$}
    \STATE $f^i \gets \text{SAM\_encoder}(I_{\text{input}}^i)$
\ENDFOR
\STATE $F_{\text{ori}} \gets \text{Stack}(f^1, \dots, f^D)$

\STATE \textbf{Restore Spatial Dimensions}
\STATE $F_{\text{up}} \gets \text{Up}(F_{\text{ori}}, (B, C, H', W', D))$
\STATE $F_{\text{general}} \gets \text{Crop}(F_{\text{up}}, (B, C, H, W, D))$

\STATE \textbf{Feature enhancement} \\
\STATE $F_{\text{final}} \gets \text{3D\_head}(F_{\text{general}})$

\STATE \textbf{return} $F_{\text{final}}$
\end{algorithmic}
\end{algorithm}

\textbf{Domain Adaptation of Foundation Models.}
After extracting structure-aware feature representation from both fixed and moving images, we develop a lightweight 3D convolutional module to generate enhanced feature representations $[C, H, W, D]$. Considering that foundation model encoders produce high-dimensional features (e.g., SAM's 256-channel output) that increase computational costs, and that conventional 2D slice-wise processing compromises volumetric consistency, 
this module uses 3D convolutions to jointly reduce computational complexity through efficient feature fusion and improve spatial continuity by enhancing inter-slice correlations, enabling better capture of medical image features for improved registration performance.
% this module employs 3D convolution operations to achieve dual optimization: (1) efficient multi-channel feature fusion to reduce computational complexity, and (2) enhanced inter-slice feature correlation to preserve spatial continuity of anatomical structures. This module enables the model to more effectively capture domain-specific features in medical images, leading to improved performance in downstream registration tasks.

The overall workflow can be formulated in Algorithm  \ref{alg:adaptation_pipeline}. With these simple techniques and domain adaptation, the visual foundation model achieves registration-specific adaptation without introducing a large number of additional parameters.

\subsection{Pyramidal Deformation Field Prediction} To handle both small and large deformations, coarse-to-fine registration based on pyramid features is an efficient scheme. Following previous research \cite{chen2024encoder}, the input feature embeddings are first downsampled hierarchically to generate multi-scale feature maps. Subsequently, the displacement field is progressively optimized at each level, from coarse to fine. In each layer, we take the moving features and fixed features as inputs and utilize a three-layer convolution block to predict the corresponding velocity field $\mathbf{u}$. After integration \cite{dalca2019unsupervised}, the deformation field $\phi^l_0$ can be obtained. If there is a lower layer, the resultant deformation fields $\phi^{l+1}$ would be upsampled to the same resolution as the current layer and composed with $\phi^l_0$ for progressive refinement, formulated as,
\begin{align}
    \tilde{\phi}^{l+1}  &= {up}(\phi^{l+1}), \nonumber &\\
    \phi^{l}  &= \tilde{\phi}^{l+1} \circ \Delta \phi^{l}_0,
\end{align}
where \({up}(\cdot)\) denotes \(2\times\) trilinear upsampling and scaling, \(\text{exp}(\cdot)\) refers to the scaling and squaring function applied to the displacement field, and \(\circ\) denotes the warping function. Note that, on the bottom layer, the deformation field $\phi^{l}$ is exactly $\phi^{l}_0$. This design captures global deformation patterns at coarse levels and optimizes local details at fine levels, thereby achieving globally precise registration. 
% The number of pyramid layers $l$ directly affects the registration accuracy, where tasks for larger deformation would need more pyramid layers. 

\subsection{Hierarchical Loss Function} 

% The loss function typically includes two terms: a dissimilarity loss and a regularization loss. The former measures the distance between the warped moving image and the fixed image at each pyramid level, while the latter ensures smooth deformation fields, avoiding unrealistic deformations. To enhance consistency, we compute losses across all pyramid layers, where moving images are downsampled and warped by the deformation field at each layer to measure dissimilarity from the downsampled fixed images.

In medical image registration, single-scale similarity metrics often struggle to achieve a balance between global alignment accuracy and the preservation of local anatomical details. Previous approaches typically compute intensity differences at the image level within a multi-scale pyramid framework, overlooking the high-level semantic consistency embedded in deep feature spaces, and critically depend on highly consistent image conditions. The potential differences in scanning conditions between moving and fixed images may result in local misalignments when using intensity-only loss functions, particularly in cases involving complex anatomical structures or images affected by noise artifacts.

To address this issue and enhance the overall alignment consistency, we introduce a hierarchical feature consistency loss, denoted as $L_{\text{HFC}}$. Specifically, the feature embeddings extracted by the SAM encoder from both the fixed and moving images are downsampled to generate multi-level feature representations, denoted as ${F}_{\text{fix}}$ and ${F}_{\text{mov}}$, respectively. At each resolution level $l$, the moving image features are first spatially transformed using the corresponding deformation field $\Phi^l$, and then the similarity discrepancy is computed with respect to the fixed image features at the same scale. The formulation of the hierarchical feature consistency loss is given as follows:
\begin{equation}
\label{eqn:hsr}
\begin{split}
{L}_{{HFC}} = \sum_{l=1}^n \frac{1}{2^{l-1}}\| F_{{mov}}^l \circ \mathbf{\phi}^{l} - F_{{fix}}^l \|.
\end{split}
\end{equation}
%where, $u^{l+1}$ is the upsampled deformation field optimized from the previous level, $\phi_{l}\left ( \cdot  \right )$ denotes the deformation operation, $L$ represents the number of layers in the feature pyramid. In our experiments, we set $n=5$.

\subsection{Overall Losses } 
Following \cite{chen2024textscf,Cheng_2025_CVPR}, SAMIR uses a loss with a dissimilarity term and a regularization term.
The dissimilarity loss combines image-level normalized cross-correlation (NCC) $L_{\text{NCC}}$ and feature-level hierarchical feature consistency loss $L_{\text{HFC}}$.
% The former measures the distance between the warped moving and the fixed images at each pyramid level, while the latter enforces smoothness of the deformation field to avoid unrealistic transformations. The dissimilarity loss captures differences at both the image level and feature level. At the image level, we employ normalized cross-correlation (NCC), denoted as $L_{\text{NCC}}$, to evaluate the similarity between the warped moving and the fixed images. At the feature level, we incorporate the hierarchical feature consistency loss $L_{\text{HFC}}$ to further enhance registration consistency. 
For the weakly supervised version, where segmentation masks are used during network training, we also compute regional dissimilarity using the Dice loss $L_{\text{Dice}}$, as proposed in \cite{chen2021deepDiscontinuity}.
\begin{equation}
\label{eqn:sim}
\begin{split}
{L}_{sim} = \lambda_0 {L}_{NCC} + \lambda_1 L_{HFC}+\lambda_2 L_{Dice},
% {L}_{sim} = \lambda_0 \times {L}_{NCC} + \lambda_1 \times L_{HFC}+\lambda_2 \times L_{Dice},
\end{split}
\end{equation}

Similar to previous works \cite{dalca2019unsupervised}, we adopt a diffusion-based smoothness regularization term $L_{{smooth}}=\sum_{x \in \Omega} \|\nabla \varphi(x)\|^2$ to enforce spatial smoothness of the deformation field. 

% The complete loss function used for training the network is therefore formulated as,
The complete loss function is formulated as,
\begin{equation}
\label{eqn:total}
\begin{split}
{L}_{total} = {L}_{sim} +\lambda_3 L_{smooth}.
% {L}_{total} = {L}_{sim} +\lambda_3 \times L_{smooth},
\end{split}
\end{equation}
where, $\lambda_0$, $\lambda_1$, $\lambda_2$ and $\lambda_3$ are hyperparameters to balance different losses.

\section{Experiment}

\subsection{Experiment Setting}

\subsubsection{Datasets and Implementation Details.} We evaluate our SAMIR on two datasets: the ACDC cardiac MRI dataset \cite{bernard2018deep} and Abdomen CT dataset \cite{xu2016evaluation}. The ACDC dataset focuses on intra-subject motion tracking between end-diastole (ED) and end-systole (ES) phases, comprising 80 training, 20 validation, and 50 test cases. Bidirectional registration (ED-to-ES and ES-to-ED) yields 160, 40, and 100 image pairs for training, validation, and testing, respectively. All images were preprocessed to $128\times128\times16$ (resolution $1.8\times1.8\times10 mm^3$). The Abdomen CT dataset, from the Learn2Reg challenge, addresses inter-subject registration with large deformations across abdominal organs ($e.g.$, liver, kidneys, spleen, pancreas). It includes 30 CT scans: 20 training, 3 validation, and 7 test cases. Pairwise combinations created 380 training, 6 validation, and 42 test pairs, processed to $192\times160\times256$. SAM features were precomputed to avoid redundant extraction. The experiments utilized the Adam optimizer with a learning rate of $1e^{-4}$. All experiments were implemented in PyTorch and executed on a single NVIDIA RTX A6000 GPU. The hyper-parameters $\lambda_0-\lambda_3$ are all set to 1. 

\begin{table*}[t]
\begin{center}
{
\begin{tabular}{ lcccccc }
\hline
\hline
%\rowcolor{lightgray}
Model & Dice (\%) $\uparrow$ & HD95 (mm) $\downarrow$ & SDlogJ $\downarrow$ & MAs (G) $\downarrow$ & PS (MB) $\downarrow$ & Time $\downarrow$\\ 
\hline
Initial & 58.14 & 11.95 & - & - & - & -\\
\hline
VoxelMorph \cite{balakrishnan2018unsupervised} & 75.26 & 9.33 & 0.044 & 19.5 & 0.32 & 0.18\\
TransMorph \cite{chen2022transmorph} & 74.97 & 9.44 & 0.045 & 50.20 & 46.69 & 0.26\\
LKU-Net \cite{jia2022u} & 76.53 & 9.13 & 0.049 & 160.50 & 33.35 & 0.22\\
Fourier-Net \cite{jia2023fourier} & 76.61 & 9.25 & 0.047 & 86.07 & 17.43 & 0.27\\
\hline
CorrMLP \cite{meng2024correlation} & 77.31 & 9.00 & 0.056 & 47.59 & 4.19 & 0.28\\
MemWarp \cite{zhang2024memwarp} & 76.74 & 9.67 & 0.108 & 1270.00 & 47.78 & 0.58\\
RDP \cite{wang_RDP} & 78.06 & 9.02 & 0.076 & 154.00 & 8.92 & 0.36\\
\hline
\textbf{SAMIR-vith (Ours)} & \textbf{80.74} & 8.22 & 0.048 & 230.34 & 7.15 & 0.32\\
\hline
\hline
\end{tabular}
}
\end{center}
\caption{
Quantitative comparison on the cardiac ACDC dataset. 
 Statistically significant improvements in registration accuracy are highlighted in bold.
% Best-performing metrics are highlighted in bold. 
Symbols indicate direction: $\uparrow$ for higher is better, $\downarrow$ for lower is better. 
``Initial" refers to baseline results before registration.
}\label{tab:ACDC}
\end{table*}

\begin{table}[tb]
\begin{center}
{
\setlength{\tabcolsep}{2pt}
\small % 调整字体大小：\tiny, \scriptsize, \footnotesize, \small等
\begin{tabular}{ lccc }
\hline
\hline
%\rowcolor{lightgray}
Model &  Dice (\%) $\uparrow$ & HD95 (mm) $\downarrow$ & SDlogJ $\downarrow$ \\ 
\hline
Initial & 30.68 & 29.77 & - \\
% ConvexAdam~\cite{siebert2021fast} & 84.64 & 1.50 & \textbf{0.07} \\
\hline
VoxelMorph  & 47.05 & 23.08 & 0.13 \\
TransMorph  & 47.94 & 21.53 & 0.13 \\
LKUNet  & 52.78 & 20.56 & 0.98 \\
\hline
LapIRN  & 54.55 & 20.52 & 1.73 \\
CorrMLP  & 56.11 & 19.52 & 0.16 \\
RDP  & 58.77 & 20.07 & 0.22 \\
MemWarp  & 60.24 & 19.84 & 0.53 \\
\hline
%FourierNet \cite{jia2023fourier} & 48.49 & 20.69 & 0.13 \\
FourierNet  & 42.80 & 22.95 & 0.13 \\
\hline
ConvexAdam  & 51.10 & 23.14 & 0.11\\ 
SAMConvex  & 53.65 & 18.66 & 0.12 \\
\hline
\textbf{SAMIR-vith (Ours)} & \bf{66.68} & \bf{13.45} & 0.17 \\
\hline
\hline
\end{tabular}
}
\end{center}
%\vspace{-2ex}
\caption{
Quantitative comparison on abdominal CT dataset. 
% Quantitative comparison on the abdomen CT dataset. 
% Statistically significant better results are highlighted in bold.
Statistically significant results are presented in bold.
% , with symbols following the same convention as in Table \ref{tab:ACDC}.
% Best-performing metrics are in bold. 
% Symbols indicate direction: $\uparrow$ for higher is better, $\downarrow$ for lower is better. 
% "Initial" refers to baseline results before registration. 
}
\label{tab:abdomen}
\end{table}

\subsubsection{Comparison Methods and Evaluation Metrics.} We compared our method with SOTA deformable image registration approaches, including VoxelMorph\cite{balakrishnan2018unsupervised}, TransMorph\cite{chen2022transmorph}, LKUNet\cite{jia2022u}, Fourier-Net\cite{jia2023fourier}, CorrMLP\cite{meng2024correlation}, MemWarp \cite{zhang2024memwarp}, and RDP\cite{wang_RDP}, where the latter three using feature pyramid architectures. For both ACDC and abdominal datasets, we used publicly available codes and fine-tuned each model for optimal performance. For the abdominal dataset, we additionally evaluated LapIRN \cite{2020LaplacianPyramidNetworks}(incompatible with short-axis data), ConvexAdam\cite{siebert2021fast}, and SAMConvex\cite{li2023samconvex} (effective for large deformations).Following \cite{dalca2019unsupervised,chen2022transmorph,zhang2024memwarp,chen2024textscf}, we employed the Dice Similarity Coefficient (Dice) and the 95\% Hausdorff Distance (HD95) to assess anatomical alignment, and standard deviation of the Jacobian determinant logarithm (SDlogJ) to evaluate the smoothness of deformation fields. Computational efficiency was measured by multi-adds (MAs), parameter size (PS), and average inference time.

\subsection{Results and Analysis} 

\subsubsection{Intra-subject Registration on ACDC.}
% We validate our method for intra-subject registration using cardiac MRI images from ACDC. 
As shown in Table \ref{tab:ACDC}, the pyramid-based methods (CorrMLP, MemWarp, RDP) consistently outperform single-layer deformation approaches, including traditional convolution (VoxelMorph), transformer-based (TransMorph), large-kernel convolution (LKU-Net), and Fourier-Net architectures. This highlights the effectiveness of pyramid-based methods in capturing both small and large deformations. With robust structure-aware feature learning from SAM, SAMIR significantly outperforms those pyramid-based registration methods ($p<0.05$), with a 2.68\% improvement over the sub-optimal method RDP. 
Note that SAM features can be precomputed, requiring only ~0.94 seconds per ACDC sample for extraction, and thus impose no additional computational burden during registration.
% Note that the SAM feature can be computed prior to the network training and testing, and therefore, the incorporation of the SAM feature would not enlarge the realistic computation burden on the registration process, as demonstrated in the table.

\begin{figure*}[tb]
\begin{center}
\includegraphics[width=1\textwidth]{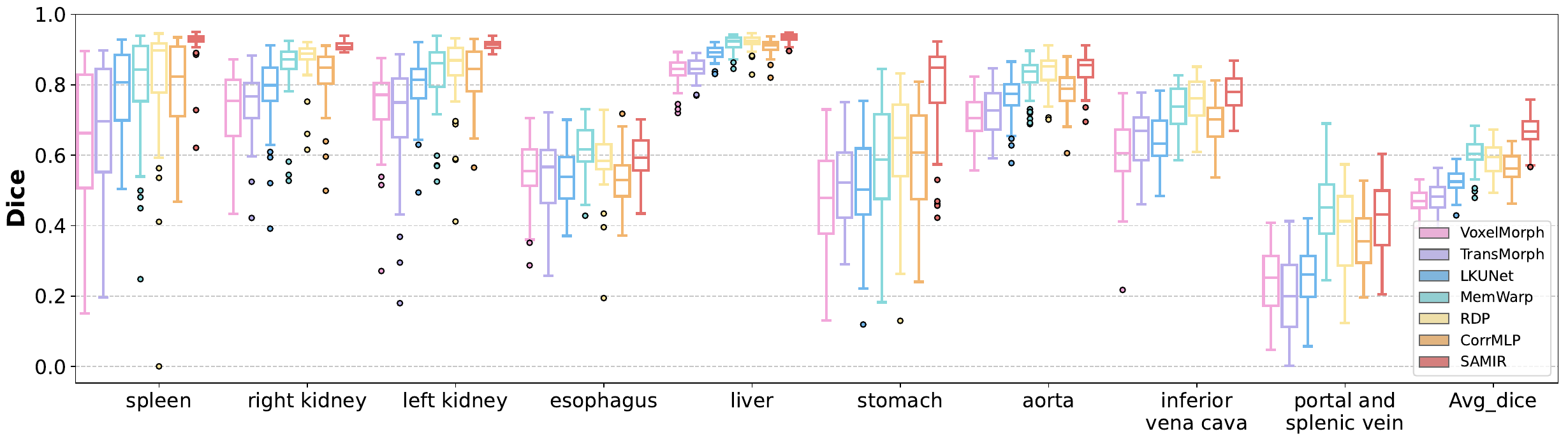}
\caption{Boxplot on the abdomen CT dataset. Our SAMIR significantly outperforms the rest approaches on all Dice scores.} %Each colored box represents the distribution of data for a specific method, with the horizontal line inside the box indicating the median. The upper and lower boundaries of the box correspond to the 25th and 75th percentiles, respectively. The height of the box reflects the interquartile range, illustrating the dispersion of the data. Circles in the plot represent outliers. In the box plot, red denotes the proposed SAMIR method.
\label{fig:boxplot}
\end{center}
\end{figure*}

\begin{figure*}[tb]
\begin{center}
\includegraphics[width=1\textwidth]{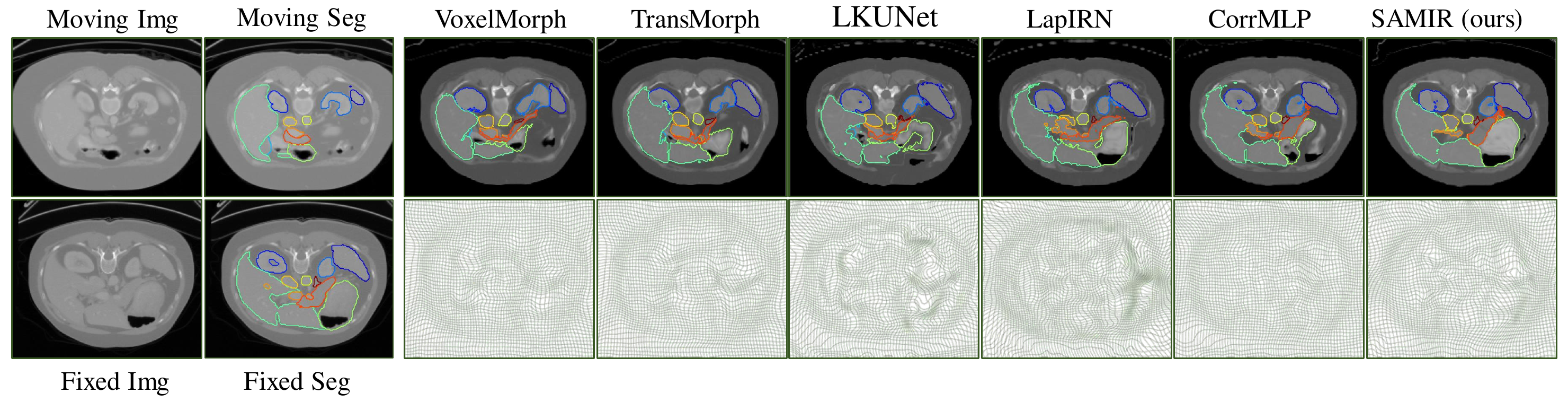}
\caption{Visual comparison between our SAMIR and SOTA methods on the abdomen CT dataset. %Left column: Moving and fixed images; Right column: corresponding warped moving image (first row), deformation fields (second row).
}
\label{fig:contract}
\end{center}
\end{figure*}

\subsubsection{Inter-subject Registration on Abdomen Dataset.} We demonstrated SAMIR's effectiveness in large deformation tasks on abdominal CT, as shown in Tabla \ref{tab:abdomen} and Figure \ref{fig:boxplot}.
% We demonstrated the effectiveness of SAMIR in handling large deformation scenarios on the abdominal CT dataset, as shown in Tabla \ref{tab:abdomen} and Figure \ref{fig:boxplot}.
Similar to the results on ACDC, the pyramid-based registration methods significantly outperform VoxelMorph, TransMorph, and LKUNet, as the deformation across different subjects is larger than intra-subject registration. Our SAMIR achieves significantly better registration performance than them, with a 6.44\% improvement on the average Dice score than the sub-optimal approach MemWarp. 
The boxplot shows SAMIR outperforms traditional methods in overall and per-structure Dice scores.
% The boxplot demonstrates that our SAMIR outperforms traditional approaches on both the average Dice score and the Dice score of each anatomical structure.

\subsection{Ablation Study} 
To evaluate the contribution of each module, we conducted an ablation study on the abdominal dataset, and the results are shown in Table \ref{tab:ablation_abdomen}. 
% Two variants were created: Model1 (without the SAM-based encoder) and Model2 (without the HFC loss), with results shown in Table \ref{tab:ablation_abdomen}. 
% Compared to SAMIR, Model1 shows a 1.78\% drop in average Dice, demonstrating the SAM encoder's capability to capture semantic consistency across anatomical structures by generating structure-aware visual features. Removing $L_{HFC}$ (Model2) reduces SAMIR's DSC by 0.49\%, demonstrating that the HFC loss's multi-scale feature alignment can further enhance registration accuracy.
The results show that both the SAM encoder and HFC loss enhance registration performance. This outcome confirms the effectiveness of SAM's structure-aware visual features and validates the multi-scale feature alignment capability of the HFC loss.
% (detailed table in Appendix)
% general-purpose 

\begin{table}[tb]
\begin{center}
{
\setlength{\tabcolsep}{3pt}
\small
\begin{tabular}{ ccccc }
\hline
\hline
\multicolumn{2}{c}{Modules} & \multirow{2}{*}{Dice (\%) $\uparrow$} & \multirow{2}{*}{HD95(mm)$\downarrow$} & \multirow{2}{*}{SDlogJ $\downarrow$} \\ 
\cline{1-2} 
 SAM encoder & HFC loss & & & \\
\hline
   & $\checkmark$ & 64.90 & 15.86 & 0.17 \\
 $\checkmark$ &   & 66.19 & 13.93 & 0.17   \\
 $\checkmark$ & $\checkmark$ & \bf{66.68} & 13.45 & 0.17 \\
\hline
\hline
\end{tabular}
}
\end{center}
\caption{
Evaluation of the effectiveness of the SAM encoder and HFC loss used in SAMIR on the abdominal CT dataset.
}
\label{tab:ablation_abdomen}
\end{table}

% \begin{table*}[tb]
% \begin{center}
% {
% %\setlength{\tabcolsep}{4pt}
% \begin{tabular}{ lccccc }
% \hline
% \hline
% \multirow{2}{*}{Model} & \multicolumn{2}{c}{Modules} & \multirow{2}{*}{Dice (\%) $\uparrow$} & \multirow{2}{*}{HD95(mm)$\downarrow$} & \multirow{2}{*}{SDlogJ $\downarrow$} \\ 
% \cline{2-3} 
% & SAM image encoder & HFC loss & & & \\
% \hline
% Model-1 &   & $\checkmark$ & 64.90 & 15.86 & 0.17 \\
% Model-2 & $\checkmark$ &   & 66.19 & 13.93 & 0.17   \\
% \textbf{SAMIR (Ours)} & $\checkmark$ & $\checkmark$ & \bf{66.68} & \bf{13.45} & 0.17 \\
% \hline
% \hline
% \end{tabular}
% }
% \end{center}
% \caption{
% Evaluation of the effectiveness of the SAM image encoder and HFC loss used in SAMIR on the abdomen CT dataset.
% }
% \label{tab:ablation_abdomen}
% \end{table*}

\subsubsection{Different Versions of SAM.} SAM offered three backbones with varying parameter sizes: ViT-B, ViT-L, and ViT-H. To assess model scale impact, we compared these backbones on the ACDC dataset, as shown in Table \ref{tab:sam}. Experiments showed that ViT-L and ViT-H outperformed ViT-B in Dice coefficients, but they share a similar registration performance. This may be because ViT-L already captures sufficient structure information for subsequent registration, and further parameter increases provided limited benefit. %Overall, the small performance differences suggest our method achieves efficient registration even with smaller memory configurations, offering flexibility and practicality for clinical deployment.
In addition, we also compared with a fine-tuned ViT-B model released in MedSAM\cite{MedSAM}, which was trained on 1,570,263 medical image masks covering 10 imaging modalities. We observed some improvement over the original model, but the gain was limited($p>0.05$). This indicates that the 3D Head has learned sufficient medical domain characteristics of the target and adapted to the downstream task. Moreover, our framework can be easily integrated with other large models, suggesting further potential for performance improvement.

\begin{table}[tb]
\begin{center}
{
\small
\begin{tabular}{ lccc }
\hline
\hline
%\rowcolor{lightgray}
Model & Dice (\%) $\uparrow$ & HD95 (mm) $\downarrow$ & SDlogJ $\downarrow$\\ 
\hline
SAM-ViT B & 80.49 & 8.39 & 0.046 \\
SAM-ViT L & 80.74 & 8.27 & 0.047 \\
SAM-ViT H & 80.74 & 8.22 & 0.048 \\
MedSAM-ViT B & 80.97 & 7.90 & 0.073 \\
\hline
\hline
\end{tabular}
}
\end{center}
\caption{
Comparison of the performance of different SAM models on the ACDC dataset.
}
\label{tab:sam}
\end{table}

\subsubsection{Different Input Sizes for the SAM Encoder.}
% To determine the optimal value of the image size $k \times k$ input to the SAM encoder, we conducted experiments with different input sizes shown in Tab. \ref{tab:samsize_acdc} and Tab. \ref{tab:samsize_abdomen}. The results show that the model performs best when upsampled to $k=512$, which may be due to artifacts generated during the upsampling process.
To determine the optimal input image size $k \times k$ for the SAM encoder, we conducted ablation experiments under various resolution settings shown in Table \ref{tab:samsize_abdomen}. The results show that for abdominal image registration, the model achieves the best performance when the input is upsampled to $k=1024$. Given the high variability in organ size and morphology within the abdominal cavity, higher spatial resolution helps enhance the visibility of smaller structures (e.g. left and right adrenal gland), leading to more accurate alignment. However, since the performance improvement was not significant ($p > 0.05$) and the computational cost increased substantially, we ultimately selected $k = 512$.

\begin{table}[tb]
\begin{center}
{
\small
\begin{tabular}{ lccc }
\hline
\hline
%\rowcolor{lightgray}
k &  Dice (\%) $\uparrow$ & HD95 (mm) $\downarrow$ & SDlogJ $\downarrow$ \\ 
\hline
256 & 65.36 & 14.45 & 0.17  \\
512 & 66.68 & 13.45 & 0.17\\
1024 & 67.05  & 13.36 & 0.17 \\
\hline
\hline
\end{tabular}
}
\end{center}
\caption{
Comparison of dice with different input sizes for the SAM encoder on Abdomen CT dataset.
}
\label{tab:samsize_abdomen}
\end{table}

\subsection{Robustness Evaluation}
\begin{figure}[tb]
\centering
\includegraphics[width=0.9\columnwidth]{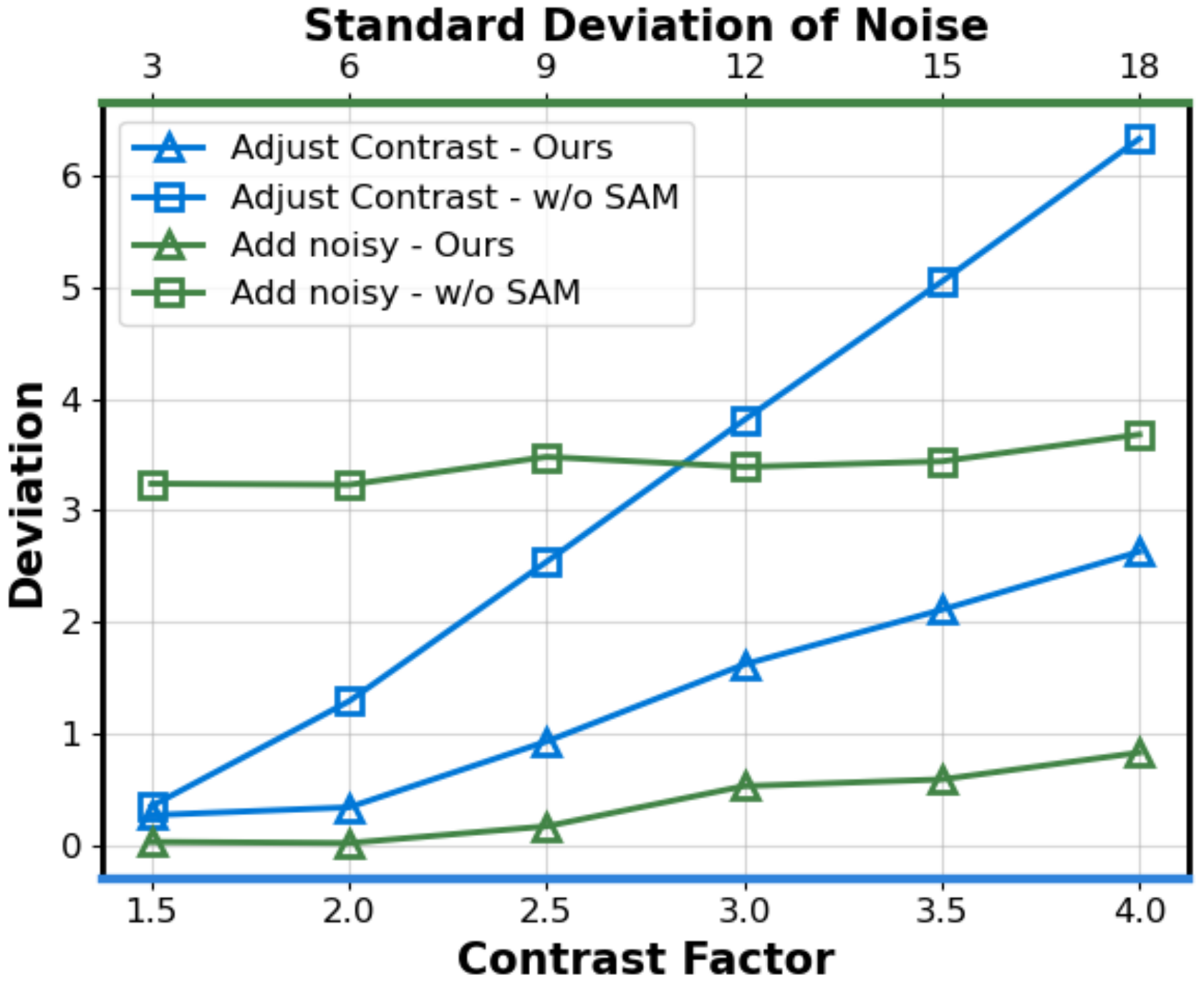} 
\caption{Registration performance of SAMIR under different interference.}
\label{fig:noisy}
\end{figure}

Our SAMIR utilizes structure-aware features rather than raw images for registration, potentially offering improved interference robustness over traditional image-only approaches. To evaluate the model's robustness, we conducted experiments by adjusting the contrast through gamma correction and adding Gaussian noise with varying standard deviations to the input data, as shown in Figure \ref{fig:noisy}. Analysis reveals that SAMIR achieves superior performance in handling noise and intensity variance across moving and fixed images, which can be attributed to the discriminative structure-aware features learned by the SAM encoder.
% Moreover, since the HFC loss aligns features at the feature level, models trained with HFC loss exhibit greater stability when faced with degraded image quality or shifts in image distribution.

% \subsection{Discussion} Although SAMIR achieves SOTA performance in both inter-subject and intra-subject registration tasks, it still has several limitations. The lightweight 3D convolutional module introduced after the SAM image encoder enhances inter-slice feature correlations, but its ability to model 3D spatial continuity is limited by module complexity, potentially resulting in sub-optimal representation of complex anatomical structures. 
% % Additionally, current experiments only validate the SAM encoder's basic capability for general-purpose visual feature extraction, which could be further improved by medical domain adaptation or joint training strategies. 
% Additionally, the present study primarily evaluates the feature extraction capability of the standard SAM encoder. Its performance could potentially be enhanced through more sophisticated medical domain adaptation techniques or integration with other advanced vision foundation models.
% Furthermore, as a modality-invariant feature, SAM features could also benefit multi-modality image registration, which could be a direction for future work.

\section{Conclusion}
This paper proposed a feature-driven registration framework, SAMIR, via efficient feature learning based on the SAM encoder. A SAFE block composing the SAM encoder is designed to extract efficient modality-invariant features from the raw images, with a PDFP block to achieve coarse-to-fine registration, in order to handle large deformation. The HFC loss is used to further enhance the anatomical structure consistency on a multi-scale. Experimental results demonstrate that SAMIR, through its efficient feature learning mechanism, achieves significantly superior registration accuracy compared to SOTA methods while exhibiting enhanced robustness against intensity variations and noise interference. Future research directions include improving inter-slice correlation in SAM feature extraction and extending the framework to multi-modality registration scenarios. %Thorough experiments demonstrate that, via efficient feature learning, our SAMIR achieved significantly better registration performance than SOTA approaches, and presents superior robustness to intensity and noise interferene. Future work would further enhance the inter-slice correlation of SAM features and demonstrate its feasibility on multi-modality registration tasks.

\appendix
% \subsection{References}
\bibliography{samir}

% \newpage
% \input{AnonymousSubmission/LaTeX/ReproducibilityChecklist}

\end{document}